\documentclass[sigconf]{acmart}

\usepackage{booktabs} 
\usepackage[capitalise]{cleveref}
\usepackage{multirow}
\usepackage{graphicx}
\usepackage{subcaption}
\usepackage{enumitem}

\usepackage{balance} 
\usepackage[american]{babel}
\usepackage{microtype}

\newcommand\newcite[1]{\citeauthor{#1}~\cite{#1}}

\setcopyright{rightsretained}

\acmConference[WISDOM'18]{the 7th KDD Workshop on Issues of Sentiment Discovery and Opinion Mining)}{August 2018}{London, UK}
\acmYear{2018}
\copyrightyear{2018}

\usepackage{microtype}
\title{Market Trend Prediction using Sentiment Analysis:\\Lessons Learned and Paths Forward}

\begin{document}


\author{Andrius Mudinas}
\affiliation{%
  \institution{Birkbeck, University of London}
  \streetaddress{Malet Street}
  \city{London WC1E 7HX}
  \country{UK}
}
\email{andrius@dcs.bbk.ac.uk}

\author{Dell Zhang}
\affiliation{%
  \institution{Birkbeck, University of London}
  \streetaddress{Malet Street}
  \city{London WC1E 7HX}
  \country{UK}
}
\email{dell.z@ieee.org}

\author{Mark Levene}
\affiliation{%
  \institution{Birkbeck, University of London}
  \streetaddress{Malet Street}
  \city{London WC1E 7HX}
  \country{UK}
}
\email{mark@dcs.bbk.ac.uk}


\begin{abstract}
Financial market forecasting is one of the most attractive practical applications of sentiment analysis.
In this paper, we investigate the potential of using sentiment \emph{attitudes} (positive vs negative) and also sentiment \emph{emotions} (joy, sadness, etc.) extracted from financial news or tweets to help predict stock price movements.
Our extensive experiments using the \emph{Granger-causality} test have revealed that (i) in general sentiment attitudes do not seem to Granger-cause stock price changes; and (ii) while on some specific occasions sentiment emotions do seem to Granger-cause stock price changes, the exhibited pattern is not universal and must be looked at on a case by case basis.
Furthermore, it has been observed that at least for certain stocks, integrating sentiment emotions as additional features into the machine learning based market trend prediction model could improve its accuracy.
\end{abstract}

\maketitle



\section{Introduction}
\label{sec:Introduction}

In recent years, a whole industry has been formed around financial market sentiment detection~\cite{Xing2018,Xing2018a}.
Traditional financial news/data providers, notably Thomson Reuters and Bloomberg, have started providing commercial sentiment analysis services.
As a result, new financial platforms, such as StockTwits\footnote{\url{http://stocktwits.com}} which offers sentiment analysis tools, have also emerged.
Nowadays, many investment banks and hedge funds are trying to exploit the sentiments of investors to help make better predictions about the financial market. 
Some of the most prominent financial institutions (including DE Shaw, Two Sigma, and Renaissance Technologies) have been reported to utilise sentiment signals, in addition to structured transactional data (like past prices, historical earnings, and dividends), in their sophisticated machine learning models for algorithmic trading.

In this paper, we aim to re-examine the application of sentiment analysis in the financial domain.
Specifically, we try to answer the following research question.
\begin{quote}
\emph{Can market sentiment really help to predict stock price movements?}
\end{quote}
Although our intuition and experience both tell us that sentiment and price are correlated, it is not clear which is the cause and which is the effect. 
Furthermore, we also have little idea of what exact types of sentiment are really relevant.

Before we embark on our investigation, it is necessary for us to clarify that in this paper, we use the term ``sentiment'' to describe all kinds of affective states~\cite{Picard1995,Wiebe1999} and we draw a distinction between sentiment \emph{attitudes} and sentiment \emph{emotions} following the typology proposed by Scherer~\cite{Scherer2000}.
By attitude, we mean the narrow sense of sentiment (as in most research papers on sentiment analysis) --- whether people are \texttt{positive} or \texttt{negative} about something.
By emotion, we mean the eight ``basic emotions'' in four opposing pairs --- \texttt{joy-sadness}, \texttt{anger-fear}, \texttt{trust-disgust}, and \texttt{anticipation-surprise}, as identified by Plutchik~\cite{Plutchik1980}.

The rest of this paper is organised as follows.
In \cref{sec:RelatedWork}, we review related research.
In \cref{sec:Datasets}, we describe the text data sources from which sentiment signals have been extracted: Financial Times (FT) news articles, Reddit WorldNews Channel (RWNC) headlines, and Twitter messages (tweets).
In \cref{sec:Causality}, we conduct the \emph{Granger-causality} test~\cite{GrangerCaus} to find out whether sentiment attitudes and sentiment emotions cause stock price changes, or is it actually the other way around.
In \cref{sec:Prediction}, we carry out extensive experiments to see if a strong baseline model that utilises fifteen technical indicators for market trend prediction can be further enhanced by adding sentiment attitude and/or sentiment emotion features.
Finally, in \cref{sec:Conclusions}, we give some concluding remarks and discuss future directions.

The source code for our implemented system is open to the research community\footnote{\url{https://github.com/AndMu/Market-Wisdom}}.

\section{Related Work}
\label{sec:RelatedWork}

The ability to predict price movements on the financial market would offer a lucrative competitive edge over other market participants.
Therefore, it is not surprising that this topic has attracted much attention from both academic researchers and industrial practitioners.
According to the \emph{efficient market hypothesis} (EMH), it is impossible to ``beat the market'', since stock market efficiency always causes existing share prices to incorporate and reflect all relevant market information.
However, many people have challenged this claim and declared that it is possible to predict price movements with more than 50\% accuracy~\cite{huang2005forecasting,qian2007stock}.

A variety of technical approaches to market trend prediction have been proposed in the research literature, ranging from AutoRegressive Integrated Moving Average (ARIMA)~\cite{wang1996stock,pai2005hybrid} to ensemble methods~\cite{qian2007stock}.
\newcite{huang2005forecasting} in their work demonstrated the superiority of Support Vector Machines (SVM) in forecasting weekly movement directions of the NIKKEI 225 index, and \newcite{lin2009short} managed to achieve 70\% accuracy by combining decision trees and neural networks.
Recent advances in \emph{deep learning} have brought a new wave of methods~\cite{chen2015lstm,gao2016stock} to this field.
In particular, the Long-Short Term Memory (LSTM) recurrent neural network has been shown to be very effective.

Numerous studies have been carried out to understand the intricate relationship between sentiment and price on the financial market.
\newcite{wang2015crowds} investigated the correlation between stock performance and user sentiment extracted from StockTwits and SeekingAlpha\footnote{\url{https://seekingalpha.com/}}.
\newcite{ding2015deep} proposed a deep learning method for event-driven stock market prediction and achieved nearly 6\% improvements on S\&P 500 index prediction.
\newcite{Arias:2014} investigated whether information extracted from Twitter can improve time series prediction, and found that indeed it could help predict the trend of volatility indices (e.g., VXO, VIX) and historic volatilities of stocks.
\newcite{twitterSenti} in their research identified that some emotion dimensions, extracted from Twitter messages, can be good market trend predictors.
Similar to our approach, \newcite{deng2011combining} combined technical analysis with sentiment analysis.
However, they only used a limited set of technical indicators together with a generic lexicon-based sentiment analysis model, and attempted to predict future prices using simple regression models.

Twitter market sentiment analysis is also related to the problem of \emph{stance detection} (SD)~\cite{sobhani2016detecting}.
As defined by \newcite{mohammad2016semeval}, a typical sentiment detection system classifies the text into positive, negative or neutral categories, while in SD the task is to detect the text that is favourable or unfavourable to a specific given target.
Most of the existing research on SD is focused on the area of politics~\cite{lai2016friends,lai2018stance,taule2017overview}.
Financial market participants also often express strong stances towards particular stocks (which can be divided into the so-called ``bulls'' and ``bears'').
However, there are non-trivial differences between political sentiment and market sentiment, as the financial market is usually more cyclical and dynamic, has different sentiment drivers, and can be impacted by various external factors (e.g., company performances and geopolitical events).
Moreover, market sentiment extracted from news articles rather than social media would exhibit different characteristics: the former is less about the authors' stance but more about the facts and the interpretation of events in a significantly richer context.

The sentiment analysis system used in our experiments is a publicly-available one called
\emph{pSenti}\footnote{\url{https://github.com/AndMu/Wikiled.Sentiment}}~\cite{Mudinas2012,Mudinas2018}
which is equipped with a pre-compiled built-in financial-domain sentiment lexicon.
When carrying out sentiment analysis experiments, we followed the procedure outlined by \newcite{Mudinas2018} in which domain-specific word embeddings would be constructed and domain-specific sentiment lexicons would be induced.
Our experiments on several publicly available datasets have confirmed that consistent with the reported results in the original paper, this approach could achieve around 80\% sentiment classification accuracy for Twitter messages and slightly higher accuracy for longer texts such as Financial Times (FT) news articles.

\section{Data}
\label{sec:Datasets}

To obtain relevant sentiment signals, we have collected three Financial Times (FT)\footnote{\url{http://www.ft.com}} datasets covering different time periods (see \cref{tab:FinancialDataset}).
In addition, we have obtained a large set of historical news headlines from Reddit's WorldNews Channel (RWNC): for each date in the time period we picked the top 25 headlines ranked by Reddit users' votes.
Moreover, we have also gathered from Twitter a large collection of financial tweets, which contain in their text one or more ``cashtags''.
A cashtag is simply a `\texttt{\$}' sign followed by a stock symbol (ticker).
For example, the cashtag for the company Apple Inc., whose ticker is \texttt{AAPL} on the stock market, would be \texttt{\$AAPL}.
Here we have only collected the tweets mentioning stocks from the S\&P 500 index.

For the stock price data, we have used the \emph{end of day} (EOD) adjusted close price according to the Dow Jones Industrial Average (DJIA) index.
In our experiments, we have focused on several representative companies, Apple (AAPL), Google (GOOGL), Hewlett-Packard (HPQ), and JPMorgan Chase \& Co. (JPM), as well as a couple of the most liquid FX currency pairs, EUR/USD and GBP/USD.
All such financial market data were acquired from public datasets published by Quandl\footnote{\url{https://www.quandl.com}}, Kaggel\footnote{\url{https://www.kaggle.com}}, and Bloomberg\footnote{\url{https://www.bloomberg.com/}}.

\begin{table}[!tb]\small
	\centering	
	\begin{tabular}{l|cc|r}
		\toprule
		Source & From & To & Count \\
		\midrule
		Financial Times I   & 2011-04-01 & 2011-12-25 &   11978 \\
		Financial Times II  & 2014-04-01 & 2014-10-26 &    9731 \\
		Financial Times III & 2014-10-26 & 2015-03-08 &    6037 \\
		Reddit              & 2008-06-08 & 2016-07-01 &   76600 \\
		Twitter             & 2014-05-01 & 2015-02-01 & 1145784 \\
		\bottomrule
	\end{tabular}
	\caption{Financial market datasets used in our experiments.}
	\label{tab:FinancialDataset}
\end{table}

\section{Causality}
\label{sec:Causality}

To verify whether market sentiments can indeed be useful for predicting stock price movements, we started the investigation with \emph{Granger-causality} test~\cite{GrangerCaus} which is a time series data-driven method for identifying causality based on a statistical hypothesis test that determines whether one time series is instrumental in forecasting the other. 
Granger-causality test has been widely accepted in econometrics as a technique to discover causality in time series data.
In the sense of Granger-causality, $x$ is a cause of $y$ if it is instrumental in forecasting $y$, where `instrumental' means that $x$ can be used to increase the accuracy of $y$'s prediction compared with considering only the past values of $y$ itself.
Essentially, Granger-causality test is a \emph{null hypothesis significance test} (NHST): the null hypothesis is that the lagged $x$-values do not explain the variation in $y$.
If the $p$-value given by the test is less than $0.10$, we would be able to reject the null hypothesis and claim that $x$ indeed Granger-causes $y$.

Through our experiments, we try to find the answers to two questions:
\emph{does market sentiment cause changes in stock price}, and conversely, \emph{does stock price cause changes in market sentiment}.

\subsection{Time Series}

\begin{figure*}[!tb]
	\centering
	\includegraphics[width=0.66\textwidth,height=0.66\textwidth]{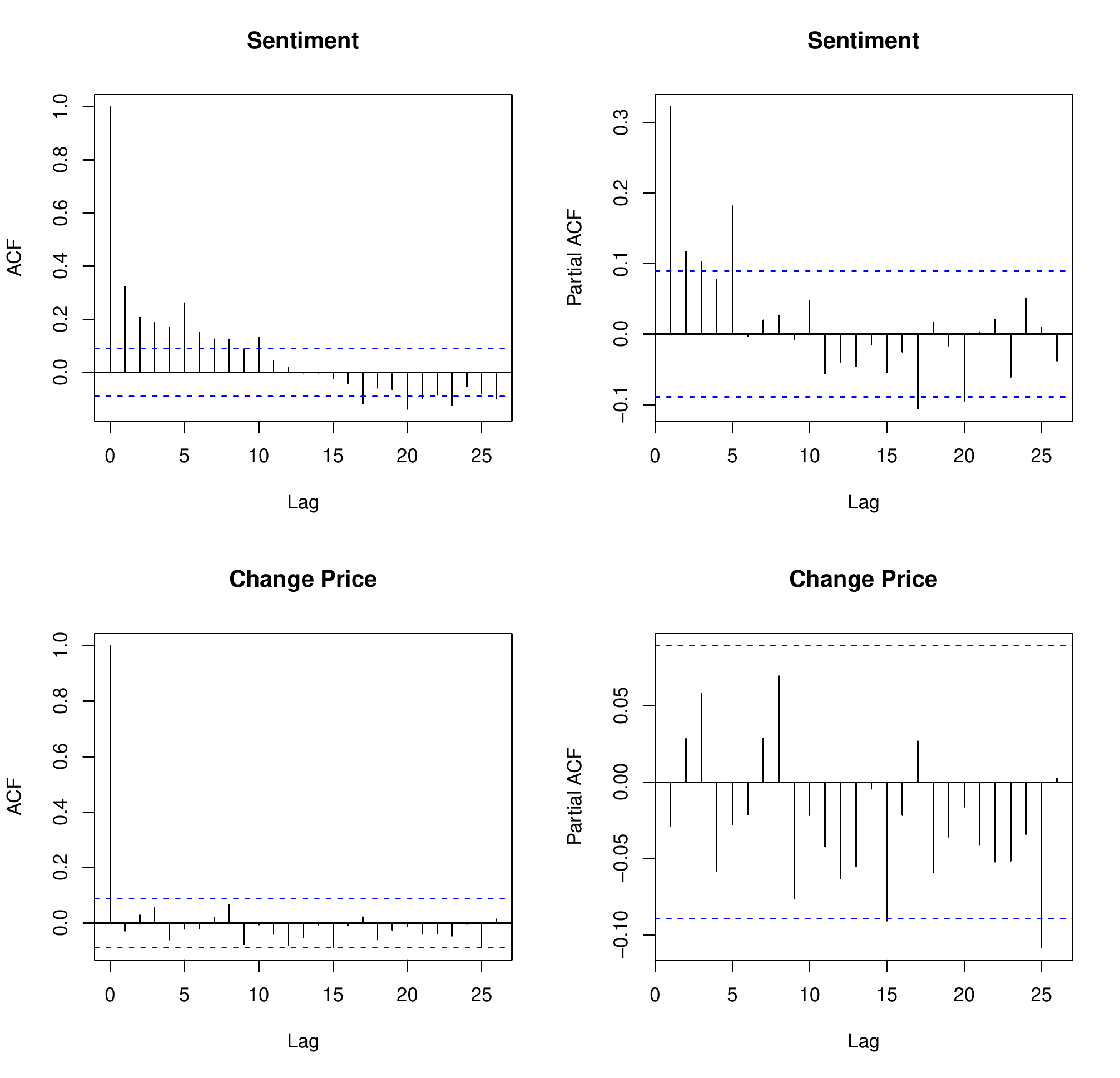}
	\caption{Stationary analysis for DJIA close prices on the FT I dataset.}
	\label{fig:Stationary}	
\end{figure*}

Before performing causality tests, it is necessary to ensure that both time series are \emph{stationary}, because otherwise the results can lead to spurious causality~\cite{he2001spurious}.
The stationarity check is typically done by analysing the autocorrelation (ACF) and partial autocorrelation (PACF) functions, and performing the Ljung-Box~\cite{ljung1978measure} or the augmented Dickey-Fuller (ADF)~\cite{dickey1979distribution} $t$-statistic tests.
Using the above methods, all our time series were verified to be stationary.
\cref{fig:Stationary} shows the stationarity check results of the FT-I and DJIA market datasets.

\begin{figure*}[!tb]
	\centering
	\begin{subfigure}[b]{0.3\textwidth}
		\includegraphics[width=\textwidth,height=\textwidth]{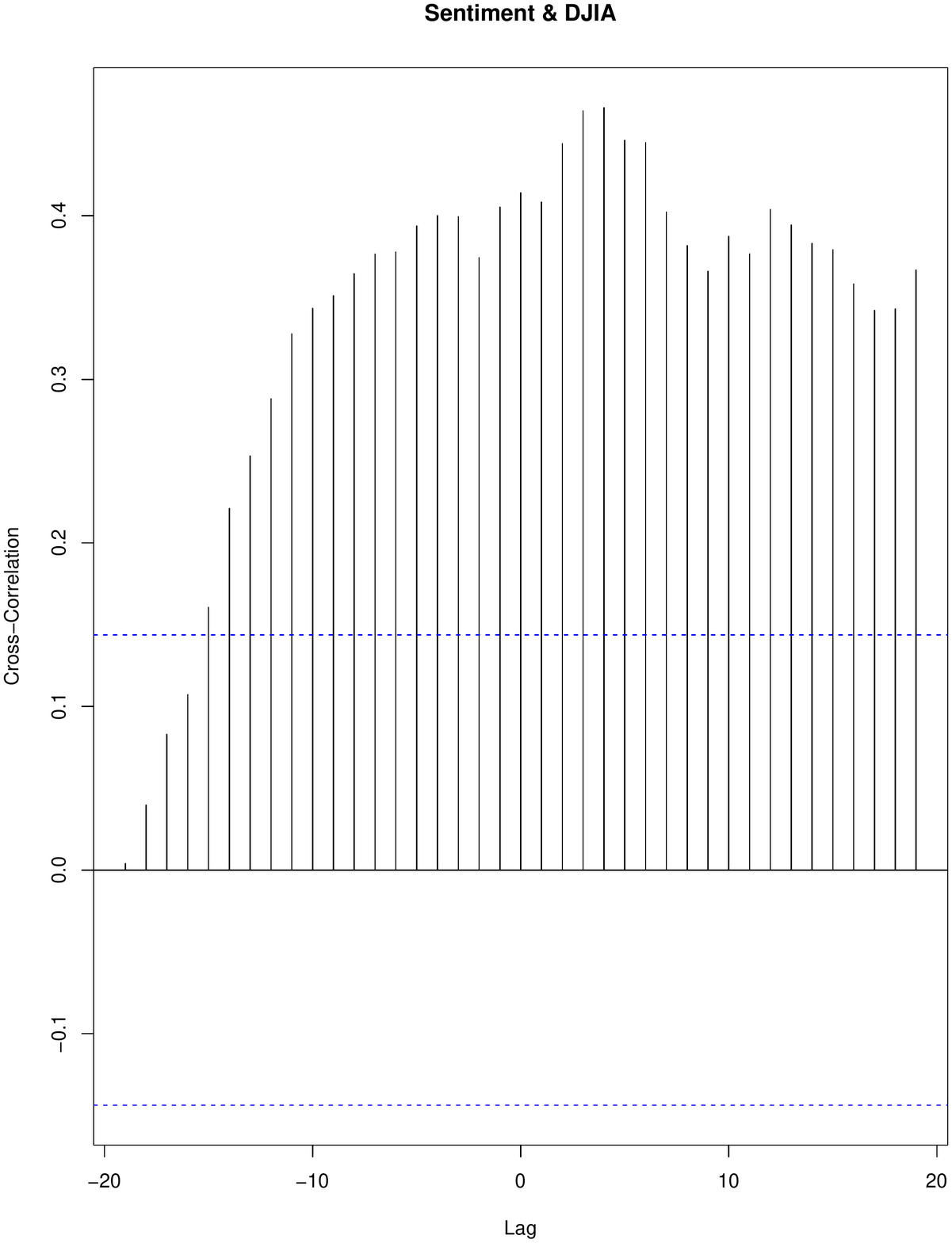}
		\caption{Financial Times I}
		\label{fig:Fin1}	
	\end{subfigure}%
	\begin{subfigure}[b]{0.3\textwidth}
		\includegraphics[width=\textwidth,height=\textwidth]{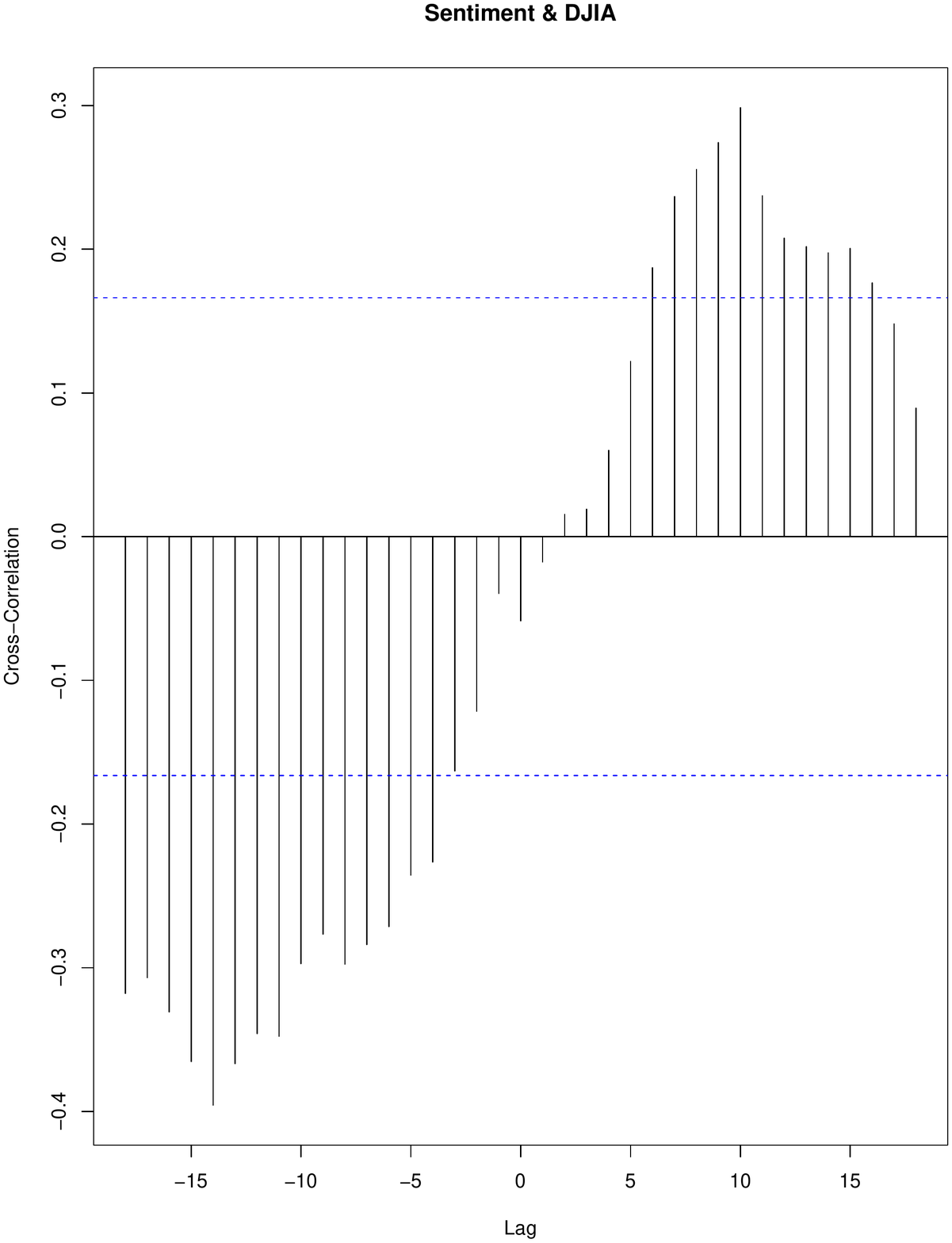}
		\caption{Financial Times II}
		\label{fig:Fin2}	
	\end{subfigure}%
	\\
	\begin{subfigure}[b]{0.3\textwidth}
		\includegraphics[width=\textwidth,height=\textwidth]{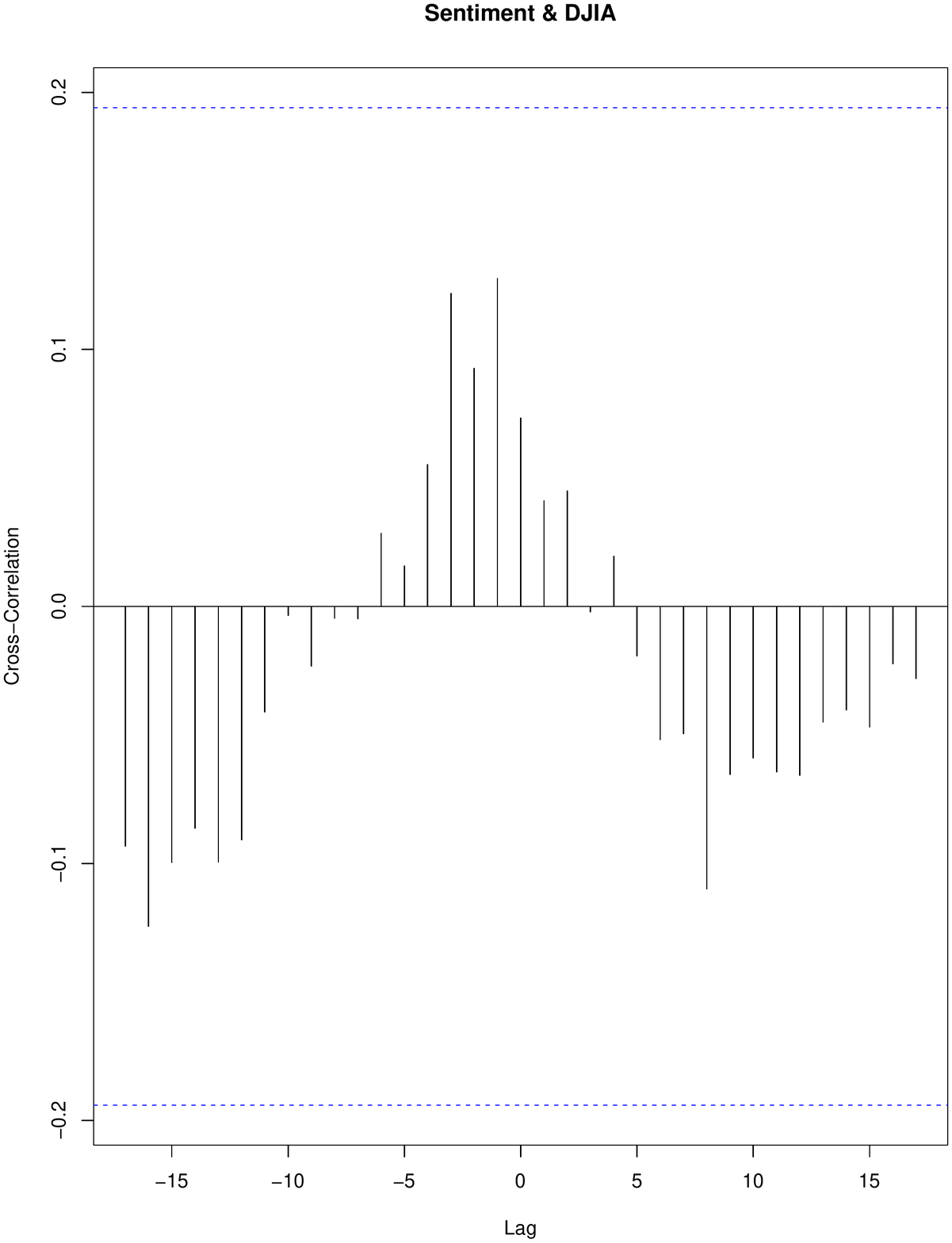}
		\caption{Financial Times III}
		\label{fig:Fin3}	
	\end{subfigure}%
	\begin{subfigure}[b]{0.3\textwidth}
		\includegraphics[width=\textwidth,height=\textwidth]{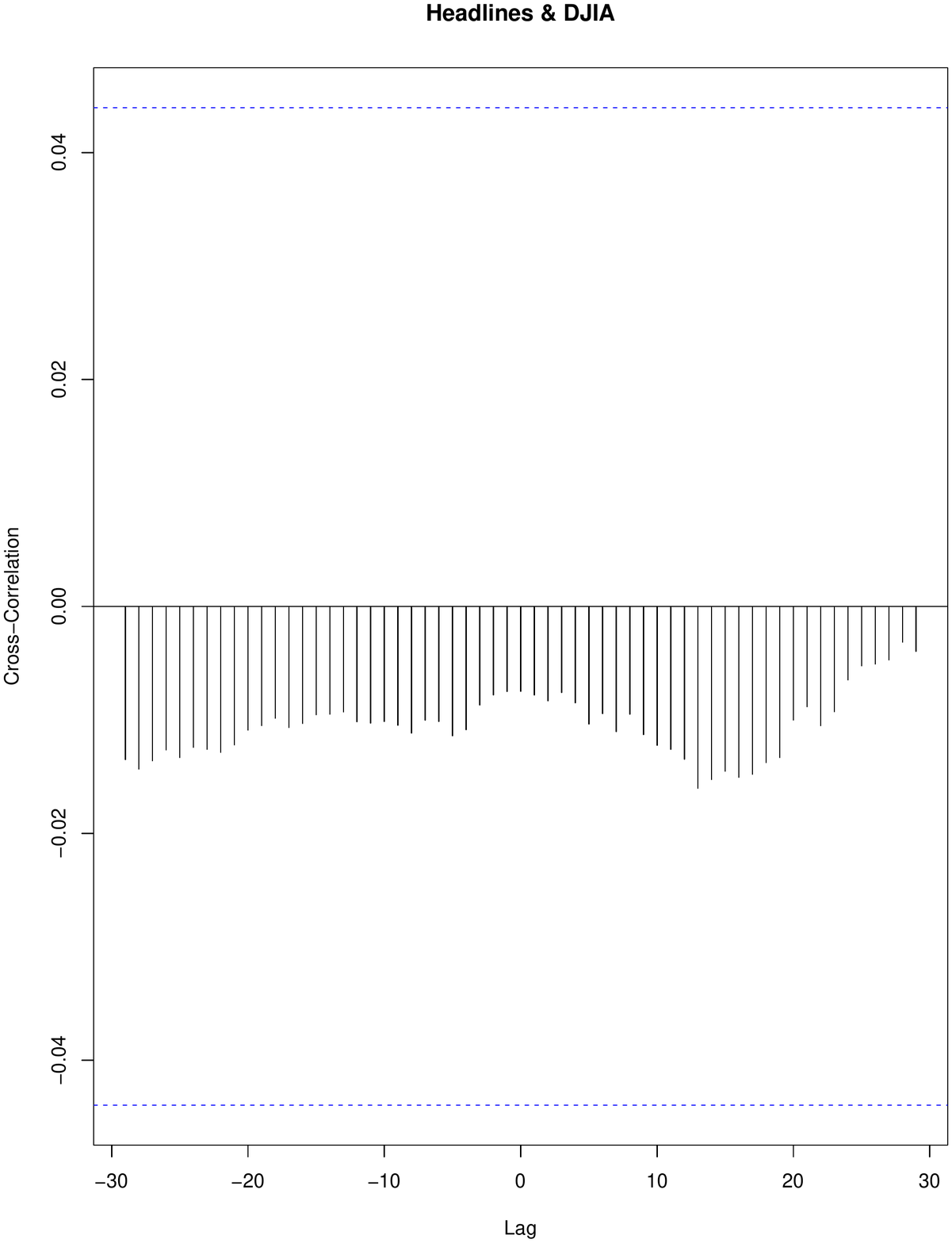}
		\caption{Reddit Headlines}
		\label{fig:Head}	
	\end{subfigure}
	\caption{The cross-correlation between sentiment attitudes and S\&P 500 prices.}
	\label{fig:Sp500Cross}
\end{figure*}

\begin{figure*}[!tb]
	\centering
	\includegraphics[width=0.66\textwidth]{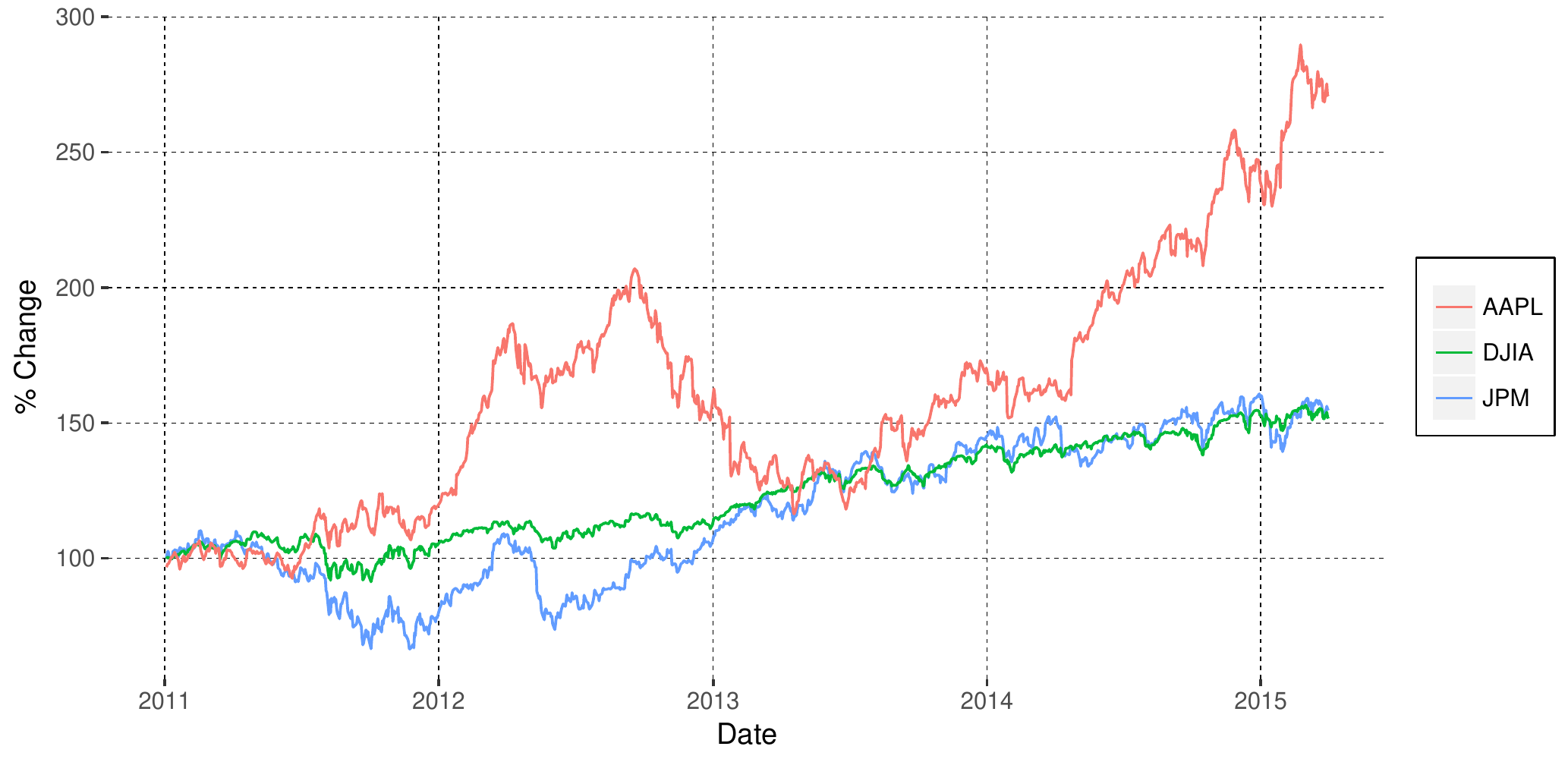}
	\caption{The market close price changes (\%).}
	\label{fig:Index}
\end{figure*}

The next step in our investigation into the relationship between sentiment and price time series is to look at their \emph{cross-correlation function} (CCF).
Although ``correlation does not imply causation'', it is frequently used as a test to discover possible causal relationship from data.
In \cref{fig:Sp500Cross} we present the cross-correlation analysis results for the S\&P 500 index on all three FT news and RWNC headlines datasets.
In the first dataset (see \cref{fig:Fin1}) we have a strong CCF between sentiment attitudes and stock prices, in the second (see \cref{fig:Fin2}) the CCF is significant in the lower left and upper right quadrants.
However, in the FT III (see \cref{fig:Fin3}) dataset, the CCF is not significant (below the confidence threshold).
This seems to suggest that the relationship between market sentiments and stock prices can be quite complex and may exist only in some of the time periods.
It is unsurprising that the financial market exhibited different behaviours in different time periods.
As shown in \cref{fig:Index}, from 2011 to mid-2013 we had a volatile market without a clear trend, whereas from 2013 until 2015 we saw a strong \emph{bull} run with continual rising prices.
Then we calculated the CCF between the sentiment attitudes found in RWNC headlines and the index prices for a longer time span from 2008 to 2016 (see \cref{fig:Head}), but still could not detect any long-term correlation.

\begin{table}[!tb]\small
	\centering
	\begin{tabular}{l|c|ccc}
		\toprule
		\multirow{2}{*}{Stock} & \multirow{2}{*}{Model} & \multirow{2}{*}{Lag} 
		    & Attitude & Price$\Rightarrow$ \\
		& & & $\Rightarrow$Price & Attitude \\
		\midrule
		\multirow{4}{*}{S\&P 500}
		& \multirow{2}{*}{Standard}
		  & 1 & 0.1929  & 0.1105 \\
		& & 2 & 0.2611  & \textbf{0.0780} \\
		\cline{2-5}
		& \multirow{2}{*}{Temporal}
		  & 1 & 0.2689  & \textbf{0.0495} \\
		& & 2 & 0.1692  & \textbf{0.0940} \\
		\hline
		\multirow{4}{*}{APPL}
		& \multirow{2}{*}{Standard}
		  & 1 & 0.7351  & 0.4253 \\
		& & 2 & 0.9117  & 0.6426 \\
		\cline{2-5}
		& \multirow{2}{*}{Temporal}
		  & 1 & 0.9478  & 0.6725 \\
		& & 2 & 0.9715  & 0.8245 \\
		\hline
		\multirow{4}{*}{GOOGL}
		& \multirow{2}{*}{Standard}
		  & 1 & 0.5285  & 0.4035 \\
		& & 2 & 0.8075  & \textbf{0.0418} \\
		\cline{2-5}
		& \multirow{2}{*}{Temporal}
		& 1 & 0.6920 & 0.5388 \\
		& & 2 & 0.8516  & \textbf{0.0422} \\
		\hline
		\multirow{4}{*}{HPQ}
		& \multirow{2}{*}{Standard}
		  & 1 & 0.1534  & 0.3996 \\
		& & 2 & 0.1877  & 0.5322 \\
		\cline{2-5}
		& \multirow{2}{*}{Temporal}
		  & 1 & 0.4069  & \textbf{0.0836} \\
		& & 2 & 0.5097  & 0.1180 \\
		\hline
		\multirow{4}{*}{JPM}
		& \multirow{2}{*}{Standard}
		  & 1 & 0.8991  & \textbf{0.0461} \\
		& & 2 & 0.9963  & \textbf{0.0435} \\
		\cline{2-5}
		& \multirow{2}{*}{Temporal}
		  & 1 & 0.9437  & 0.1204 \\
		& & 2 & 0.7722  & 0.2720 \\
		\bottomrule
	\end{tabular}
	\caption{Sentiment attitudes Granger-causality on the FT I dataset.}
	\label{tab:GrangeCausality}
\end{table}

\begin{figure*}[!tb]
	\centering
	\begin{subfigure}[b]{0.3\textwidth}
		\includegraphics[width=\textwidth,height=\textwidth]{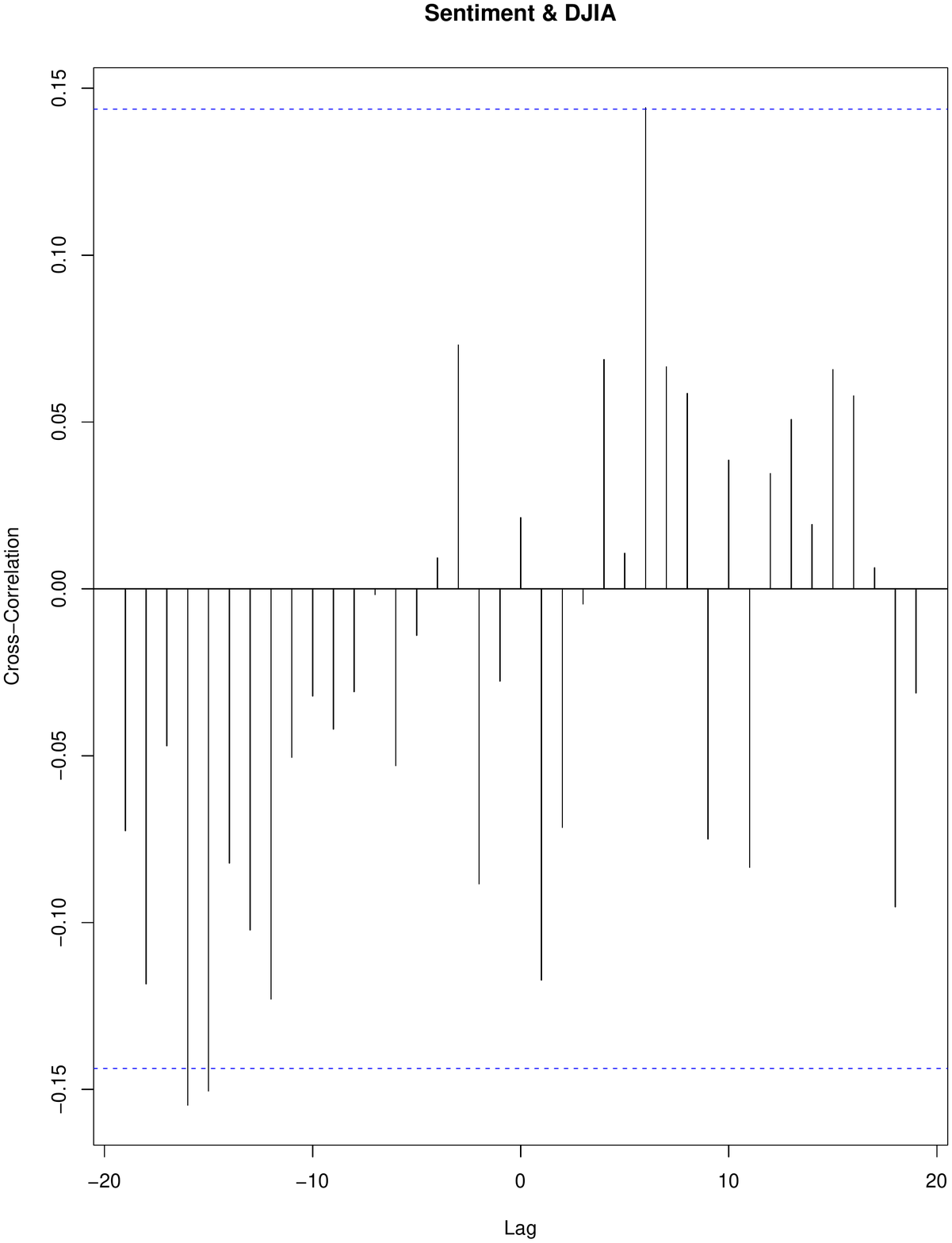}
		\caption{Financial Times I}
		\label{fig:Fin1Diff}	
	\end{subfigure}%
	\begin{subfigure}[b]{0.3\textwidth}
		\includegraphics[width=\textwidth,height=\textwidth]{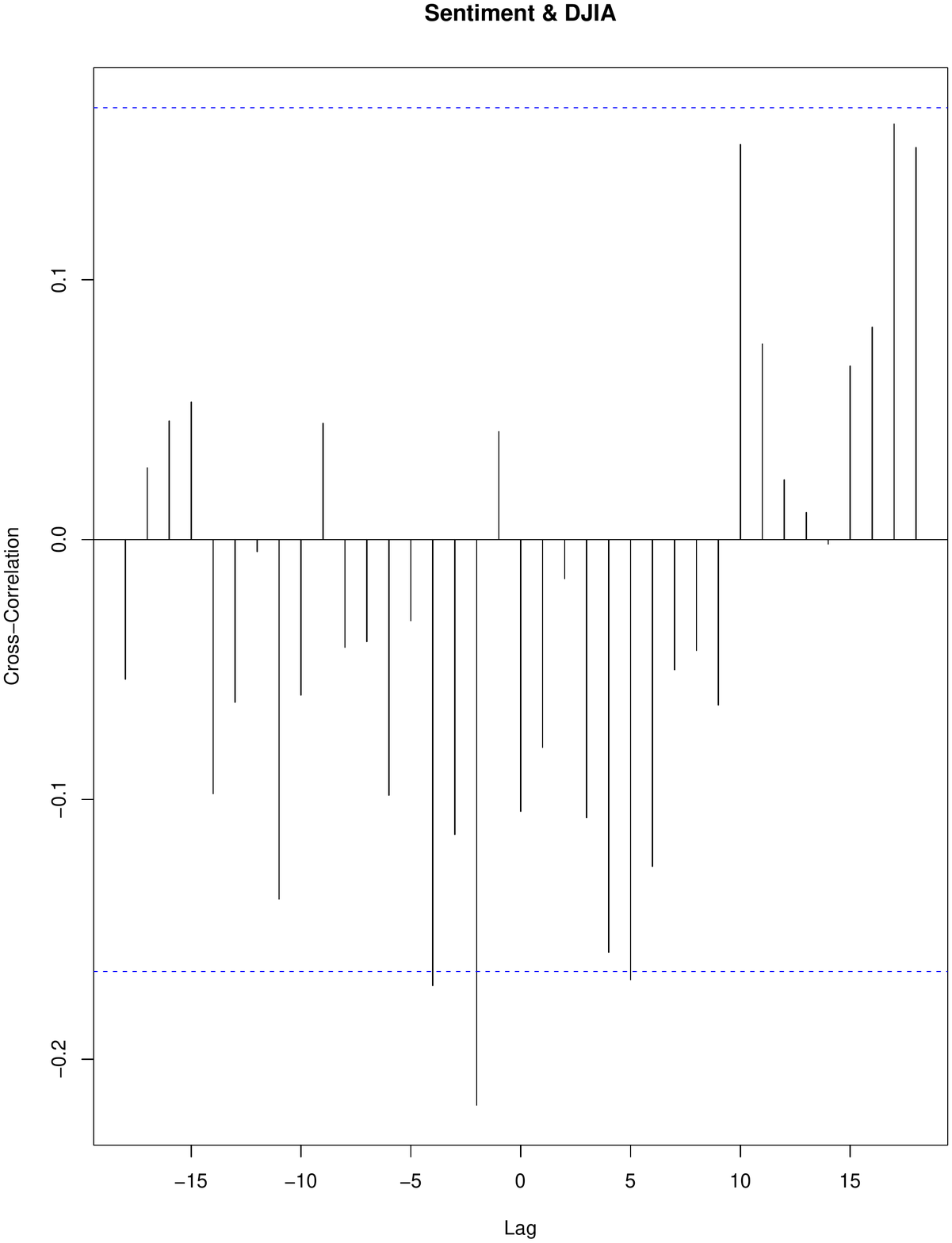}
		\caption{Financial Times II}
		\label{fig:Fin2Diff}	
	\end{subfigure}%
	\\
	\begin{subfigure}[b]{0.3\textwidth}
		\includegraphics[width=\textwidth,height=\textwidth]{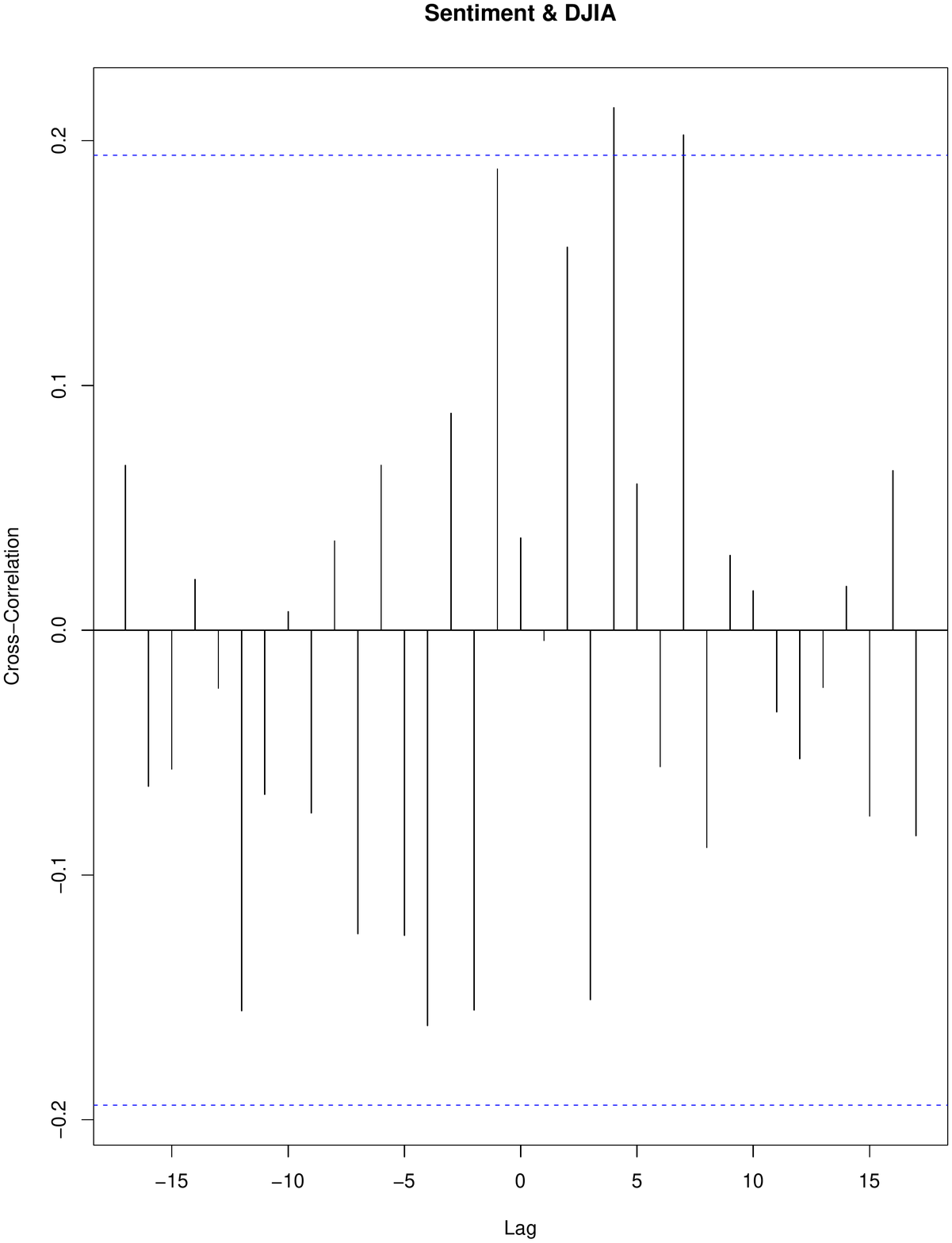}
		\caption{Financial Times III}
		\label{fig:Fin3Diff}	
	\end{subfigure}%
	\begin{subfigure}[b]{0.3\textwidth}
		\includegraphics[width=\textwidth,height=\textwidth]{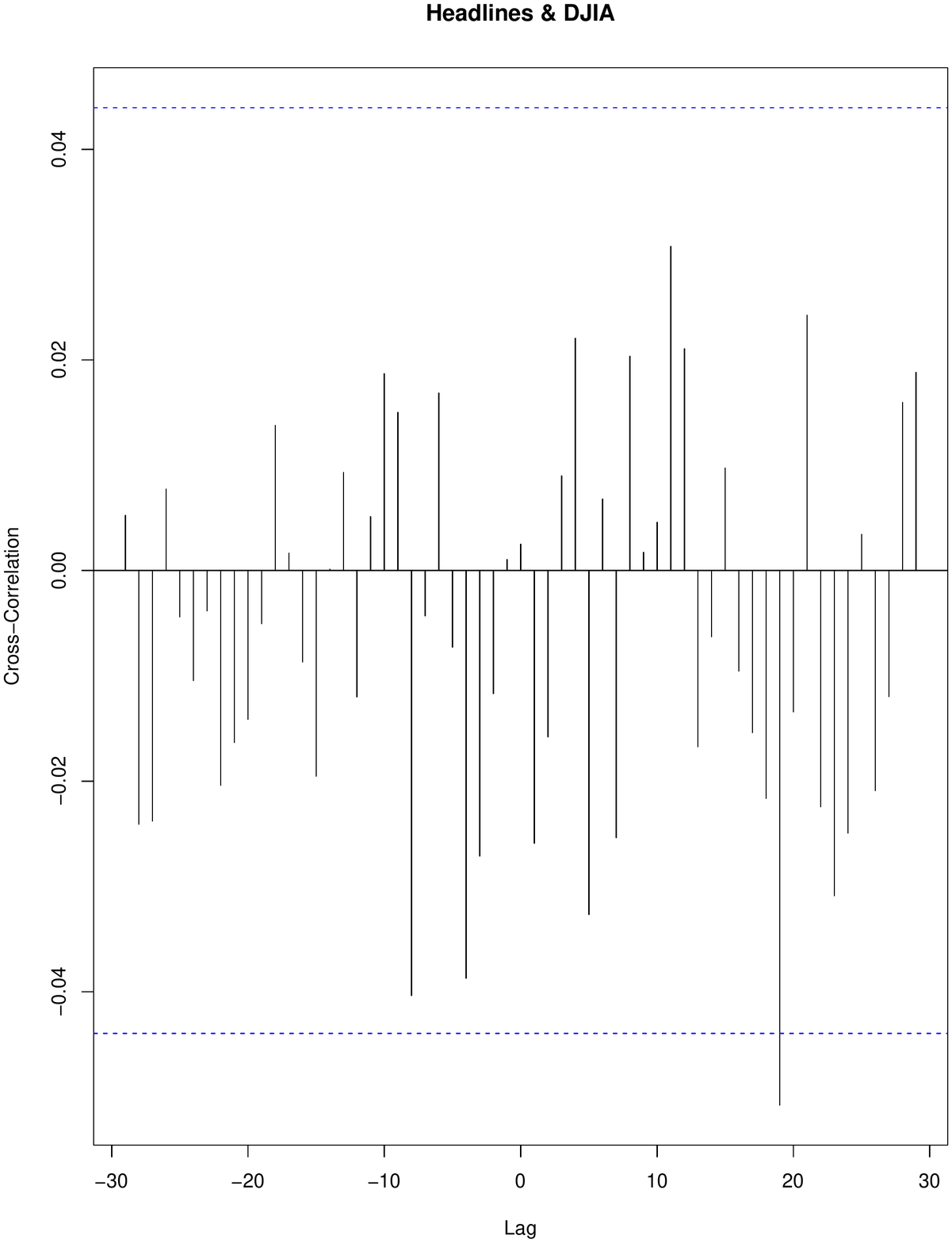}
		\caption{News Headlines}
		\label{fig:HeadDif}	
	\end{subfigure}
	\caption{The cross-correlation between sentiment attitudes and S\&P 500 price changes.}
	\label{fig:Sp500CrossDiff}
\end{figure*}

Most of the real-life automated trading systems need to make \textbf{BUY} or \textbf{SELL} decisions for the given stocks.
Therefore, from the trading perspective, the actual price of a stock is less crucial, and the profit relies on the price changes (often measured in percentages).
Similar to the previous experiments on the cross-correlation between sentiment attitudes and stock prices (see \cref{fig:Sp500Cross}), additional experiments on the cross-correlation between sentiments and stock price changes were performed (see \cref{fig:Sp500CrossDiff}).
Contrary to the previous results, we found correlations in all the FT news articles (see \cref{fig:Fin1Diff,fig:Fin2Diff,fig:Fin3Diff}) and the RWNC headlines (see \cref{fig:HeadDif}).
This suggests that the percentage changes of stock prices would have higher predictability than stock prices themselves.
However, the correlations are often only present with a substantial lag.
Therefore it is still valid that sentiment attitudes are unlikely to be useful in market trend prediction.

\subsection{Experimental Setup}

To further analyse the relationship, we performed a set of \emph{Granger-causality} tests for the S\&P 500 index and four selected stocks using the FT I dataset, in which we detected the strongest correlation.
The lags in the causality tests were set to just one or two days, considering that the financial market usually reacts to relevant news events almost instantaneously.

The sentiment analysis was performed using both a standard model and an enhanced temporal model.
In the latter, we associate the sentiments with the corresponding temporal orientations by labelling each sentence with one of the four temporal-categories (past/present/future/unknown) and calculate the sentiment strength accordingly.
Thus, the \emph{pSenti} sentiment outputs would be generalised from a single dimension into one of those four temporal-categories.
The determination of temporal orientation could be done by two different methods.
One method is to identify temporal expressions using the \emph{SuTime} temporal tagger~\cite{SuTime}.
\emph{SuTime} is a rule-based tagger built on regular expression patterns to recognise and normalise temporal expressions in the form of TIMEX3 in English text .
TIMEX3 is part of the TimeML annotation language~\cite{pustejovsky2003timebank} for marking up events, times, and their temporal relations in documents.
It recognises both absolute time (such as ``January 12, 2000'') and relative time (such as ``next month'').
The relative time can be then transformed into the corresponding absolute time using the underlying document creation time.
For example, in the sentence ``I hope next year they will release an improved version'', we would identify ``improved'' as a positive sentiment attitude with ``next year'' as its time point, so the text would lead to a positive \emph{future} sentiment feature.
Another method, which we also employed in the temporal model, is to use the tense of the sentence as the clue to determine which temporal category should be assigned to the sentiment found in the sentence.
Intuitively, only the sentiments about the present and the future value of the stock would have significant impacts on its price.
Therefore, we would filter out all sentiment scores with the \emph{past} tag.

In each of our causality test experiments, two competing hypotheses would be examined: market sentiments cause stock price changes and vice versa.

\subsection{Experimental Results}

The results obtained from the experiments (see \cref{tab:GrangeCausality}) show a mixed picture.
In all the experiments, we failed to discover any sign that sentiment attitudes Granger-cause stock price changes, which would suggest that in general sentiment attitudes probably cannot be useful for the prediction of stock price movements.
However, in many cases, we found that the opposite was true --- stock price changes Granger-cause sentiment attitudes in the news, with the strongest causality found using the temporal sentiment analysis model.
The individual stocks also produced mixed results, with each company behaving differently.
For the Apple stock, we failed to detect any causality.
For the Google stock, we identified that the prices would Granger-cause sentiment attitudes, but only with a two-day lag.
For the HP stock, we detected causality only in temporal sentiment and only with a one-day lag.
For the JPM stock, we found causality using standard sentiment, but it was absent using temporal sentiment.
It is difficult to draw general conclusions from such varying results. 
According to the Granger-causality test with a one-day or two-day lag, sentiment attitudes do not seem to be useful for predicting stock price movements.
However, the opposite seems to be true --- the sentiment attitudes should be predicted using stock price movements.
It is still possible that the Granger-causality from sentiment attitudes to stock price changes is present at a finer time granularity (e.g., minutes), but we are unable to perform such an analysis using our current datasets.

\begin{table}[!tb]\small
	\centering
	\begin{tabular}{l|ccccc}
		\toprule
		\multirow{3}{*}{Emotion} & \multirow{3}{*}{Lag} & \multicolumn{2}{c}{Standard} & \multicolumn{2}{c}{Temporal} \\ 
		& & Emotion & Price$\Rightarrow$ & Emotion & Price$\Rightarrow$ \\
		& & $\Rightarrow$Price & Emotion & $\Rightarrow$Price & Emotion \\		
		\midrule
		\multirow{2}{*}{anger} 
		& 1 & 0.3815 & 0.6299 & 0.2555 & 0.4155 \\
		& 2 & 0.3402 & 0.9153 & 0.3097 & 0.6886 \\
		\hline
		\multirow{2}{*}{anticipation} 
		& 1 & 0.5320 & 0.2650 & 0.9216 & 0.9389 \\
		& 2 & 0.4989 & 0.5765 & 0.4930 & 0.7173 \\
		\hline
		\multirow{2}{*}{disgust} 
		& 1 & 0.6668 & \textbf{0.0688} & 0.2482 & 0.2031 \\
		& 2 & 0.7166 & 0.3118 & 0.1160  & 0.2852 \\
		\hline
		\multirow{2}{*}{fear}  
		& 1 & 0.5821 & 0.1255 & 0.8698 & \textbf{0.0591} \\
		& 2 & 0.8934 & 0.2601 & 0.9888 & 0.1604 \\
		\hline
		\multirow{2}{*}{joy} 
		& 1 & 0.6972 & 0.5549 & 0.3521 & 0.1530 \\
		& 2 & 0.5567 & 0.8451 & 0.4045 & 0.4089 \\
		\hline
		\multirow{2}{*}{sadness} 
		& 1 & 0.3885 & 0.1067 & \textbf{0.0258} & 0.1019 \\
		& 2 & 0.6166 & 0.2027 & \textbf{0.0983} & 0.1423 \\
		\hline
		\multirow{2}{*}{surprise}  
		& 1 & 0.5866 & 0.7022 & 0.3830 & 0.2315 \\
		& 2 & 0.9802 & 0.8414 & 0.8445 & 0.3838 \\
		\hline
		\multirow{2}{*}{trust} 
		& 1 & 0.9983 & 0.6892 & 0.9490 & 0.1124 \\
		& 2 & 0.5534 & 0.8523 & 0.9586 & 0.2239 \\
		\bottomrule
	\end{tabular}
	\caption{Sentiment emotions Granger-causality: S\&P 500.} 
	\label{tab:GrangerMulti}
\end{table}

\newcite{twitterSenti} attempted to predict the behaviour of the stock market by measuring the sentiment emotion of people on Twitter and identified that some of the emotion dimensions have predictive power.
To verify their findings, we employed a similar model based on Plutchik's emotion dimensions extracted using the NRC sentiment lexicon~\cite{Mohammad13} and \emph{pSenti}.
In the S\&P 500 index analysis (see \cref{tab:GrangerMulti}), we found that only \texttt{sadness} could Granger-cause stock price changes, which is different from the results of \newcite{twitterSenti}.
Such a discrepancy might be explained by the fact that \newcite{twitterSenti} used different emotion dimensions, lexicons, and a different time period in their analysis.
An interesting finding which we have obtained from the experimental results is that some individual stocks like HP (see \cref{tab:grangerHP}) and JPM (see \cref{tab:grangerJPM}) have significantly more emotion dimensions with predictive power than others.
It could be seen in those cases that some emotion dimension other than \texttt{sadness}, including \texttt{surprise}, \texttt{fear}, \texttt{joy} and \texttt{trust}, also demonstrated predictive power.
On both Google (see \cref{tab:grangerGoogle}) and Apple (see \cref{tab:grangerApple}) stock price data, we failed to find any emotion causality on their stock price.
These results indicate that even if in some cases there is substantial Granger-causality from sentiment emotions to stock price changes, it is not a general pattern, and should be looked at on a case by case basis.
To find out why that is happening, it would be necessary to perform further investigation, which is beyond the scope of this paper.

\begin{table}[!tb]\small
	\centering
	\begin{tabular}{l|ccccc}
		\toprule
		\multirow{3}{*}{Emotion} & \multirow{3}{*}{Lag} & \multicolumn{2}{c}{Standard} & \multicolumn{2}{c}{Temporal} \\ 
		& & Emotion & Price$\Rightarrow$ & Emotion & Price$\Rightarrow$ \\
		& & $\Rightarrow$Price & Emotion & $\Rightarrow$Price & Emotion \\		
		\midrule
		\multirow{2}{*}{anger}
		& 1 & 0.9452 & 0.3512 & 0.6490 & 0.2352 \\
		& 2 & 0.9851 & 0.4367 & 0.7703 & 0.1461 \\
		\hline
		\multirow{2}{*}{anticipation}
		& 1 & 0.5237 & 0.8272 & 0.3032 & 0.1245 \\
		& 2 & 0.6368 & 0.3331 & 0.595  & 0.1518 \\
		\hline
		\multirow{2}{*}{disgust}
		& 1 & 0.2412 & 0.4128 & 0.1376 & 0.9851 \\
		& 2 & 0.5154 & 0.5877 & 0.3392 & 0.3130 \\
		\hline
		\multirow{2}{*}{fear}
		& 1 & 0.3717 & \textbf{0.0867} & 0.2727 & 0.5577 \\
		& 2 & 0.5609 & 0.1698 & 0.4139 & 0.1114 \\
		\hline
		\multirow{2}{*}{joy}
		& 1 & 0.3301 & 0.9946 & 0.7916 & 0.2580 \\
		& 2 & 0.6657 & 0.5264 & 0.9843 & 0.4312 \\
		\hline
		\multirow{2}{*}{sadness}
		& 1 & 0.2217 & 0.8139 & 0.2280 & 0.6620 \\
		& 2 & 0.1669 & 0.4266 & 0.1710 & 0.3245 \\
		\hline
		\multirow{2}{*}{surprise}
		& 1 & 0.9413 & 0.1960 & 0.1083 & 0.6093 \\
		& 2 & 0.9733 & 0.2433 & 0.2033 & 0.2609 \\
		\hline
		\multirow{2}{*}{trust}
		& 1 & 0.5663 & 0.8439 & 0.3219 & 0.3539 \\
		& 2 & 0.8760 & 0.5520 & 0.4473 & 0.2608 \\
		\bottomrule
	\end{tabular}
	\caption{Sentiment emotions Granger-causality: APPL.}
	\label{tab:grangerApple}
\end{table}

\begin{table}[!tb]\small
	\centering
	\begin{tabular}{l|ccccc}
		\toprule
		\multirow{3}{*}{Emotion} & \multirow{3}{*}{Lag} & \multicolumn{2}{c}{Standard} & \multicolumn{2}{c}{Temporal} \\ 
		& & Emotion & Price$\Rightarrow$ & Emotion & Price$\Rightarrow$ \\
		& & $\Rightarrow$Price & Emotion & $\Rightarrow$Price & Emotion \\		
		\midrule
		\multirow{2}{*}{anger}
		& 1 & 0.2706 & 0.4460 & 0.4420 & 0.1530 \\
		& 2 & 0.1709 & 0.7454 & 0.2457 & 0.2677 \\
		\hline
		\multirow{2}{*}{anticipation}
		& 1 & 0.1137 & 0.1951 & 0.2720 & 0.4348 \\
		& 2 & 0.1487 & 0.4839 & 0.3363 & 0.7986 \\
		\hline
		\multirow{2}{*}{disgust}
		& 1 & 0.4459 & 0.3250 & 0.4000 & 0.5865 \\
		& 2 & 0.7031 & 0.4294 & 0.6608 & 0.8880 \\
		\hline
		\multirow{2}{*}{fear}
		& 1 & 0.3362 & 0.1020 & 0.2874 & 0.1211 \\
		& 2 & 0.2763 & \textbf{0.0757} & 0.3011 & 0.1765 \\
		\hline
		\multirow{2}{*}{joy}
		& 1 & 0.4718 & \textbf{0.0417} & 0.8350 & \textbf{0.0959} \\
		& 2 & 0.7855 & \textbf{0.0998} & 0.9755 & 0.2282 \\
		\hline
		\multirow{2}{*}{sadness}
		& 1 & 0.4184 & 0.1316 & 0.3917 & 0.1782 \\
		& 2 & 0.6599 & 0.1236 & 0.5286 & 0.3844 \\
		\hline
		\multirow{2}{*}{surprise}
		& 1 & 0.6551 & \textbf{0.0606} & 0.6869 & \textbf{0.0755} \\
		& 2 & 0.7604 & 0.1166 & 0.6626 & 0.2156 \\
		\hline
		\multirow{2}{*}{trust}
		& 1 & 0.5008 & \textbf{0.0727} & 0.7541 & \textbf{0.0680} \\
		& 2 & 0.5991 & 0.1302 & 0.8334 & 0.1052 \\
		\bottomrule
	\end{tabular}
	\caption{Sentiment emotions Granger-causality: GOOGL.}
	\label{tab:grangerGoogle}
\end{table}

\begin{table}[!tb]\small
	\centering
	\begin{tabular}{l|ccccc}
		\toprule
		\multirow{3}{*}{Emotion} & \multirow{3}{*}{Lag} & \multicolumn{2}{c}{Standard} & \multicolumn{2}{c}{Temporal} \\ 
		& & Emotion & Price$\Rightarrow$ & Emotion & Price$\Rightarrow$ \\
		& & $\Rightarrow$Price & Emotion & $\Rightarrow$Price & Emotion \\		
		\midrule
		\multirow{2}{*}{anger}
		& 1 & 0.2129 & 0.8639 & 0.1300 & 0.9466 \\
		& 2 & 0.4084 & 0.9521 & 0.2234 & 0.9689 \\
		\hline
		\multirow{2}{*}{anticipation}
		& 1 & \textbf{0.0757} & 0.6288 & 0.1316 & 0.7853 \\
		& 2 & 0.2279 & 0.9059 & 0.3371 & 0.8986 \\
		\hline
		\multirow{2}{*}{disgust}
		& 1 & 0.4868 & 0.8126 & 0.2001 & 0.4536 \\
		& 2 & 0.3803 & 0.9353 & 0.2252 & 0.6722 \\
		\hline
		\multirow{2}{*}{fear}
		& 1 & 0.2679 & 0.4841 & 0.1214 & 0.8193 \\
		& 2 & 0.5361 & 0.4741 & 0.2371 & 0.9319 \\
		\hline
		\multirow{2}{*}{joy}
		& 1 & \textbf{0.0399} & 0.8186 & \textbf{0.0902} & 0.6261 \\
		& 2 & 0.1255 & 0.8945 & 0.2410 & 0.7273 \\
		\hline
		\multirow{2}{*}{Sadness}
		& 1 & \textbf{0.0106} & 0.8669 & \textbf{0.0110} & 0.9208 \\
		& 2 & \textbf{0.0416} & 0.9365 & \textbf{0.0388} & 0.8456 \\
		\hline
		\multirow{2}{*}{surprise}
		& 1 & \textbf{0.0217} & 0.6825 & \textbf{0.0010} & 0.3890 \\
		& 2 & \textbf{0.0759} & 0.7830 & \textbf{0.0064} & 0.3034 \\
		\hline
		\multirow{2}{*}{trust}
		& 1 & \textbf{0.0447} & 0.8620 & \textbf{0.0766} & 0.7693 \\
		& 2 & 0.1340 & 0.9034 & 0.2158 & 0.7948 \\
		\bottomrule
	\end{tabular}
	\caption{Sentiment emotions Granger-causality: HPQ.}
	\label{tab:grangerHP}
\end{table}

\begin{table}[!tb]\small
	\centering
	\begin{tabular}{l|ccccc}
		\toprule	
		\multirow{3}{*}{Emotion} & \multirow{3}{*}{Lag} & \multicolumn{2}{c}{Standard} & \multicolumn{2}{c}{Temporal} \\ 
		& & Emotion & Price$\Rightarrow$ & Emotion & Price$\Rightarrow$ \\
		& & $\Rightarrow$Price & Emotion & $\Rightarrow$Price & Emotion \\		
		\midrule
		\multirow{2}{*}{anger}
		& 1 & 0.1788 & 0.0488 & 0.1796 & 0.2349 \\
		& 2 & 0.4903 & 0.1223 & 0.3155 & 0.3713 \\
		\hline
		\multirow{2}{*}{anticipation}
		& 1 & 0.3893 & 0.1360 & 0.1389 & 0.5989 \\
		& 2 & 0.7729 & 0.3145 & 0.3389 & 0.4732 \\
		\hline
		\multirow{2}{*}{disgust}
		& 1 & 0.2168 & 0.1260 & 0.2208 & 0.2267 \\
		& 2 & 0.5297 & 0.2913 & 0.3611 & 0.1637 \\
		\hline
		\multirow{2}{*}{fear}
		& 1 & \textbf{0.0298} & 0.1565 & \textbf{0.0214} & 0.2072 \\
		& 2 & 0.1173 & 0.2495 & \textbf{0.0210} & 0.1187 \\
		\hline
		\multirow{2}{*}{joy}
		& 1 & 0.3417 & 0.2169 & 0.1073 & 0.9574 \\
		& 2 & 0.8544 & 0.3905 & 0.3293 & 0.9086 \\
		\hline
		\multirow{2}{*}{sadness}
		& 1 & 0.6079 & 0.3038 & 0.3985 & 0.5781 \\
		& 2 & 0.9297 & 0.5856 & 0.4495 & 0.4194 \\
		\hline
		\multirow{2}{*}{surprise}
		& 1 & 0.1351 & \textbf{0.0303} & \textbf{0.0498} & 0.1145 \\
		& 2 & 0.4296 & \textbf{0.0593} & \textbf{0.0850} & \textbf{0.0461} \\
		\hline
		\multirow{2}{*}{trust}
		& 1 & \textbf{0.0991} & 0.2218 & \textbf{0.0458} & 0.6664 \\
		& 2 & 0.1232 & 0.2066 & \textbf{0.0165} & 0.6645 \\
		\bottomrule
	\end{tabular}
	\caption{Sentiment emotions Granger-causality: JPM}
	\label{tab:grangerJPM}
\end{table}

\section{Prediction}
\label{sec:Prediction}

The causality analysis in \cref{sec:Causality} has revealed that in some cases sentiment emotions could be good indicators for stock price changes.
In the next set of experiments, we would like to investigate how sentiment attitudes and/or sentiment emotions could be exploited in a machine learning model for market trend prediction to improve its accuracy.

Basically, there are two types of stock market analysis: fundamental and technical.
The former evaluates a stock based on its corresponding company' business performance, whereas the latter evaluates a stock based on its volume and price on the financial market as measured by a number of so-called \emph{technical indicators}.
Both types of analysis generate trading signals, which would be monitored by human traders or automated trading systems who use that information to execute trades.
In our experiments, only technical analysis has been utilised. 
It is likely that incorporating fundamental analysis and employing more technical indicators would improve the predictive model's performance.
However, our research objective is not to create the optimal market trend prediction system, but to analyse and understand the predictive power of sentiments on the financial market.
For this purpose, a baseline model with several common technical indicators should be good enough.

\subsection{Baseline}
\label{sec:Baseline}

We first built a baseline machine learning model to predict stock price changes with a number of selected technical indicators, and then tried to incorporate additional sentiment-based features, i.e., sentiment attitudes and sentiment emotions.

In order to construct a decent baseline model, we made use of ten common technical indicators which led to a total of fifteen features as follows.
\begin{itemize}[leftmargin=*]
	\item \underline{Moving Averages (MA)}.
	A moving average is frequently defined as a support or resistance level.
	Many basic trading strategies are centred around breaking support and resistance levels.
	In a rising market, a 50-day, 100-day or 200-day moving average may act as a support level, and in a falling market as resistance.
	We calculated 50-day, 100-day and 200-day moving averages and included each of them as a feature.
	\item \underline{Williams \%R}. 
	This indicator was proposed by Larry Williams to detect when a stock was overbought or oversold.
	It tells us how the current price compares to the highest price over the past period (10 days).
	\item \underline{Momentum (MOM)}.
	This indicator measures how the price changed over the last $N$ trading days.
	We used two momentum-based features, one-day momentum and five-day momentum.
	\item \underline{Relative Strength Index (RSI)}.
	This is yet another indicator to find overbought and oversold stocks.
	It compares the magnitude of gains and losses over a specified period.
	We used the most common 14 days period.
	\item \underline{Moving Average Convergence Divergence (MACD)}.
	This is one of the most effective momentum indicators, which shows the relationship between two moving averages.
	It generates three features: MACD, signal, and histogram values.
	\item \underline{Bollinger Bands} 
	is one of the most widely used technical indicators.
	It was developed and introduced in the 1980s by the famous technical trader John Bollinger.
	It represents two standard deviations away from a simple moving average, and can thus help price pattern recognition.
	\item \underline{Commodity Channel Index (CCI)} 
	is another a momentum indicator introduced by Donald Lambert in 1980.
	This indicator can help to identify a new trend or warn of extreme conditions by detecting overbought and oversold stocks.
	Its normal movement is in the range from -100 to +100, so going beyond this range is considered a BUY/SELL signal.
	\item \underline{Average Directional Index (ADX)} 
	is a non-directional indicator which quantifies the price trend strength using values from 0 to 100.
	It is useful for identifying strong price trends.
	\item \underline{Triple Exponential Moving Average (TEMA)} was developed by Patrick Mulloy and first published in 1994.
	It serves as a trend indicator, and in contrast to moving averages it does not have the associated lag.
	\item \underline{Average True Range (ATR)} is a non-directional volatility indicator developed by \newcite{wilder1978new}.
	The stocks and indexes with higher volatility typically have higher ATR.
\end{itemize}
The features were all normalised to zero mean and unit variance in advance.

In our context, the machine learning model is just a binary classifier that generates two kinds of signals: \textbf{BUY} ($+1$) and \textbf{SELL} ($-1$).
It aims to predict whether or not the stock's price, $n$ days in the future, will be higher ($+1$) or lower ($-1$) than today's price.
In the preliminary experiments, we tried to find out which machine learning algorithm would perform best and how far into the future the model would be able to predict.

Following the research literature in this area~\cite{huang2005forecasting,chen2015lstm,gao2016stock}, we evaluated the two most popular machine learning approaches to market trend prediction, SVM (with the RBF kernel) and LSTM recurrent neural network.
Each dataset was randomly divided into two sets: 2/3 for training and 1/3 for testing.
The parameters of the SVM and LSTM algorithms were set via grid search on the training set. 
The final LSTM model consists of a single LSTM layer with 400 units and utilises a drop-out rate of 0.5~\cite{Wager2013,Srivastava2014}.

It is common for such market trend prediction models to use a time lag of a few days and by doing so avoid short-term price volatility~\cite{das2012support}.
In our experiments, we tried both three-day and five day lags.
Similar to the previous studies by \newcite{cao2003support} and \newcite{thomason1999}, using five-day lags was found to be optimal.

The preliminary experimental results as shown in \cref{tab:MarketBaselineSelection} indicate that SVM outperformed LSTM on all the datasets.
The $F_1$ scores suggest that LSTM often favoured the positive class over the negative class and produced unbalanced results.
The reason could be that the size of the dataset is relatively small: there are 670 data points in the analysed time period 2011-2015.
Contrary to LSTM, SVM always yielded balanced and stable results.

\begin{table*}[!tb]\small
	\centering
	\begin{tabular}{l|c|ccc|ccc}
		\toprule
		\multirow{2}{*}{Type} & \multirow{2}{*}{Method} & \multicolumn{3}{c|}{3-day ahead} & \multicolumn{3}{c}{5-day ahead} \\
		&
		& Acc & $F_1^{\text{up}}$ & $F_1^{\text{down}}$
		& Acc & $F_1^{\text{up}}$ & $F_1^{\text{down}}$ \\
		\midrule
		\multirow{2}{*}{DJIA}
		& SVM  & 0.616 & 0.738 & 0.282 & \textbf{0.700} &\textbf{ 0.754} & \textbf{0.615} \\
		& LSTM & 0.559 & 0.706 & 0.120 & 0.585 & 0.728 & 0.127 \\
		\hline
		\multirow{2}{*}{AAPL}
		& SVM  & 0.577 & 0.676 & 0.391 & \textbf{0.685} & \textbf{0.723} & \textbf{0.634} \\
		& LSTM & 0.547 & 0.693 & 0.138 & 0.521 & 0.641 & 0.282 \\
		\hline
		\multirow{2}{*}{JPM}		
		& SVM  & 0.677 & 0.747 & 0.552 &\textbf{0.673} & \textbf{0.733} & \textbf{0.578} \\
		& LSTM & 0.541 & 0.665 & 0.269 & 0.573 & 0.676 & 0.373 \\
		\hline
		\multirow{2}{*}{EUR/USD}
		& SVM  & 0.642 & 0.607 & 0.672 & \textbf{0.671} & \textbf{0.620} & \textbf{0.710} \\
		& LSTM & 0.509 & 0.423 & 0.572 & 0.563 & 0.370 & 0.665 \\
		\hline
		\multirow{2}{*}{GBP/USD}
		& SVM  & 0.610 & 0.589 & 0.630 & \textbf{0.714} & \textbf{0.705} & \textbf{0.723} \\
		& LSTM & 0.500 & 0.604 & 0.323 & 0.633 & 0.646 & 0.618 \\
		\bottomrule
	\end{tabular}
	\caption{Market trend prediction using main technical indicators --- the baseline model.} 		
	\label{tab:MarketBaselineSelection}
\end{table*}

In the end, SVM with a five-day lag was selected as the baseline model which produced a reasonable accuracy of around 70\% and similar $F_1$ scores for both classes.

\subsection{Using Sentiment Signals in News}
\label{sec:NewsFeatures}

In the next set of experiments, we evaluated the predictive power of sentiments extracted from financial news articles/headlines.
The time granularity here is a single day, i.e., all sentiment-based features (including both attitudes and emotions) would be aggregated by calculating their daily averages.
If there was no sentiment information available on that day, the value zero would be assigned to the corresponding sentiment features.

The proposed new model consists of the same technical indicator features as in the baseline plus nine additional sentiment-based features:
\begin{itemize}[leftmargin=*]
	\item \underline{Sentiment attitudes}.
	The average daily sentiment attitudes, extracted using \emph{pSenti}, with values in the range from -1 to +1.
	\item \underline{Sentiment emotions} in eight categories: \texttt{anger}, \texttt{anticipation}, \texttt{disgust}, \texttt{fear}, \texttt{joy}, \texttt{sadness}, \texttt{surprise}, and \texttt{trust}, with values being the normalised occurrence frequency.
\end{itemize}

Sentiment attitudes and emotions were extracted from the FT news articles and the RWNC headlines in the time period from 2011 to 2015.
The experimental results as shown in \cref{tab:MarketFTPrediction} indicate that incorporating sentiment attitudes and sentiment emotions from the headlines actually had a negative impact on the predictive model's performance.
This is consistent with the previous section, in which no correlation or causality link was established between headlines sentiments and stock prices.
It might be explained by the fact that headlines are very short text snippets and therefore provide little chance for us to reliably detect sentiment attitudes and sentiment emotions.
The sentiments extracted from FT news articles painted a quite different picture.
The sentiment enriched model outperformed the baseline model in half of the scenarios: it demonstrated slightly better results for DJIA, JPM, and EUR/USD, but slightly worse results for AAPL and GBP/USD.
These experimental results are consistent with the previous section and confirm again that for some stocks sentiment emotions could be used to improve the baseline model for market trend prediction.

\begin{table*}[!tb]\small
	\centering
	\begin{tabular}{l|ccc|ccc|ccc}
		\toprule
		\multirow{2}{*}{Type} & \multicolumn{3}{c|}{Baseline} & \multicolumn{3}{c|}{Financial Times} & \multicolumn{3}{c}{Reddit Headlines} \\
		& Acc & $F_1^{\text{up}}$ & $F_1^{\text{down}}$
		& Acc & $F_1^{\text{up}}$ & $F_1^{\text{down}}$
		& Acc & $F_1^{\text{up}}$ & $F_1^{\text{down}}$ \\
		\midrule
		DJIA
		& 0.700 &\textbf{0.754} & 0.615 & \textbf{0.706} & 0.752 & \textbf{0.639} & 0.618 & 0.716 & 0.417 \\
		AAPL
		& \textbf{0.685} & \textbf{0.723} & \textbf{0.634} & 0.652 & 0.723 & 0.531 & 0.624 & 0.700 & 0.496 \\
		JPM
		& 0.673 & 0.733 & 0.578 & \textbf{0.679} & \textbf{0.739} & \textbf{0.583} & 0.615 & 0.713 & 0.415 \\
		EUR/USD
		& 0.641 & 0.715 & 0.518 & \textbf{0.653} & 0.691 & \textbf{0.605} & 0.638 & 0.684 & 0.578 \\
		GBP/USD
		& \textbf{0.714} & 0.705 & \textbf{0.723} & 0.711 & \textbf{0.708} & 0.715 & 0.615 & 0.625 & 0.605 \\
		\bottomrule
	\end{tabular}
	\caption{Market trend prediction using FT news articles and RWNC headlines (2011-2015).} 		
	\label{tab:MarketFTPrediction}
\end{table*}

\subsection{Using Sentiment Signals in Tweets}
\label{sec:TweetFeatures}

\begin{table*}[!tb]\small
	\centering
	\begin{tabular}{l|ccc|ccc|ccc|ccc}
		\toprule
		\multirow{2}{*}{Type} & \multicolumn{3}{c|}{baseline} & \multicolumn{3}{c|}{all+attitude+emotion}  & \multicolumn{3}{c|}{all+emotion} & \multicolumn{3}{c}{filtering+emotion} \\
		& Acc & $F_1^{\text{up}}$ & $F_1^{\text{down}}$
		& Acc & $F_1^{\text{up}}$ & $F_1^{\text{down}}$
		& Acc & $F_1^{\text{up}}$ & $F_1^{\text{down}}$
		& Acc & $F_1^{\text{up}}$ & $F_1^{\text{down}}$ \\
		\midrule
		DJIA
		& \textbf{0.810} & \textbf{0.854} & 0.727 & 0.810 & 0.846 & \textbf{0.750} & 0.778 & 0.829 & 0.682 & - & - & - \\
		AAPL
		& \textbf{0.889} & \textbf{0.918} & \textbf{0.829} & 0.810 & 0.860 & 0.700 & 0.794 & 0.847 & 0.683 & 0.794 & 0.831 & 0.735 \\
		JPM
		& 0.746 & 0.800 & 0.652 & 0.730 & 0.779 & 0.653 & 0.746 & 0.789 & 0.680 & \textbf{0.778} & \textbf{0.829} & \textbf{0.682}\\
		GBP/USD
		& \textbf{0.708} & \textbf{0.387} & \textbf{0.808} & 0.662 & 0.389 & 0.766 & 0.631 & 0.294 & 0.750 & - & - & -\\	
		EUR/USD
		& 0.685 & \textbf{0.627} & 0.727 & 0.685 & \textbf{0.627} & 0.727 & \textbf{0.692} & 0.626 & \textbf{0.739} & - & - & - \\
		\bottomrule
	\end{tabular}
	\caption{Market trend prediction using financial tweets from Twitter (01/04/2014 -- 01/04/2015).} 		
	\label{tab:MarketTwitter}
\end{table*}

In the last set of experiments, we created the enriched model based on sentiment attitudes and sentiment emotions extracted from financial tweets.
The time period of the Twitter messages dataset is significantly shorter: only from 2014 to 2015.
Consequently, the experiments were performed on a shorter time period with only 275 data points.
In this time period, almost all stock prices were continually rising (see \cref{fig:Index}).
Such a so-called bull run makes it even more difficult to assess a predictive model's performance, as any basic strategy like buy and hold would be a winning strategy.

Let us consider three different scenarios.
In the first scenario (``all+attitude+emotion''), both sentiment attitudes and sentiment emotions were extracted from all financial tweets and used as additional features.
This allowed us to verify how useful sentiment information is for market trend prediction.
In the second scenario (``all+emotion'') , only those eight sentiment emotions were used as additional features.
This provided an opportunity to validate the usefulness of sentiment emotions alone.
For the last scenario (``filtering+emotion''), only the Twitter messages (tweets) mentioning the company of our interest were utilised to extract sentiment emotions as additional features.

The experimental results as shown in \cref{tab:MarketTwitter} indicate that most of the time, the baseline model would actually outperform the expanded model with sentiment attitudes, sentiment emotions, or both as additional features.
Only for the JPM stock we could see noticeable performance improvements in the ``filtering+emotion'' setting.
Once again these results are consistent with the causality analysis in \cref{sec:Causality} and the market trend prediction experiments using financial news in \cref{sec:NewsFeatures} --- the JPM stock demonstrated that integrating sentiment emotions has the potential to enhance the baseline model.
Our results have also confirmed that sentiment attitudes on their own are probably not very useful for market trend prediction, but at least for some particular stocks sentiment emotions could be exploited to improve machine learning models like LSTM to get better market trend prediction.

Our findings are mostly in line with other researchers' results~\cite{twitterSenti}.
However, there are still many questions remaining unanswered in this area.

\section{Conclusions}
\label{sec:Conclusions}

In this paper, we have empirically re-examined the feasibility of applying sentiment analysis to make market trend predictions.

Our experiments investigated the causal relationship between sentiment attitude/emotion signals and stock price movements using various sentiment signal sources and different time periods.
The experimental results indicate that the interaction between sentiment and price is complex and dynamic: while some stocks in some time periods exhibited strong cross-correlation, it was absent in other cases.
We have discovered that in general sentiment attitudes do not seem to have any Granger-causality with respect to stock prices but sentiment emotions do seem to Granger-cause price changes for some particular stocks (e.g., JPM).
Furthermore, we have attempted to incorporate sentiment signals into machine learning models for market trend prediction.
Specifically, we have compared two popular machine learning approaches and finally selected SVM (with the RBF kernel) as the baseline which was trained using fifteen technical indicators and on average achieved 70\% accuracy for five-day market trend prediction.
The baseline model was then expanded using sentiment attitudes and sentiment emotions extracted from financial news or financial tweets as additional features.
In some scenarios, the proposed model outperformed the baseline model and demonstrated that sentiment emotions could be employed to help predict stock price movements though sentiment attitudes could not.
The sentiment emotions extracted form Financial Times news articles yielded better performances than those extracted from Reddit news headlines.

An important research question for future work is how to identify the stocks whose price changes are indeed predictable using sentiment emotions.
Although the Granger causality test on the historical data could find the stocks that were predictable in the past, there is no guarantee that they will continue to be predictable in the future.
It is very possible that a more sophisticated classifier for this purpose could be developed.

\begin{acks}
The Titan X Pascal GPU used for this research was kindly donated by the NVIDIA Corporation.
We thank the reviewers for their constructive and helpful comments.
We also gratefully acknowledge the support of Geek.AI for this work.
\end{acks}

\balance


\end{document}